# A SAREF-based Ontology for Distributed AI Workflows across the Edge-Fog-Cloud Continuum

Viorica Rozina Chifu[1], Tudor Cioara[1,*], Vasile Ofrim[1], Liana Toderean[1], Ionut Anghel[1], Laura Daniele[2], Cornelis Bouter[2]
[1] Technical University of Cluj-Napoca, Memorandumului Street 28, Cluj-Napoca 400114, Romania
[2] TNO for Innovation Life, Delft, Netherlands
viorica.chifu@cs.utcluj.ro, tudor.cioara@cs.utcluj.ro, vasile.ofrim@cs.utcluj.ro, liana.toderean@cs.utcluj.ro, ionut.anghel@cs.utcluj.ro
laura.daniele@tno.nl, cornelis.bouter@tno.nl

**Abstract**: Nowadays semantic models provide limited support for representing distributed AI workflows and their execution across heterogeneous edge, fog, and cloud environments. Therefore, AI processes and resources are often described using incompatible semantic representations, affecting the interoperability, orchestration, and reuse. To address these challenges, this paper proposes a SAREF-compliant ontology for representing distributed AI workflows across the edge-fog-cloud continuum. We extend the SAREF4SYST ontology with concepts for modeling AI pipelines, executable AI jobs, computational resources, deployment constraints, and communication relationships, providing a unified semantic model of both AI workflows and heterogeneous computing infrastructures. The ontology enables semantic interoperability, automated reasoning, and resource-aware orchestration of distributed AI applications while remaining fully aligned with the ETSI SAREF ecosystem. The ontology is evaluated using proof-of-concept smart grid energy services orchestration scenarios and validated using competency questions showing its ability to support AI workflow deployment, execution reasoning, and workload adaptation across heterogeneous edge, fog, and cloud environments. All competency questions were successfully validated using SPARQL querying and semantic reasoning. Experimental results demonstrate deployment success rates of 90-100% with average orchestration decision times below 80 ms across heterogeneous edge-fog-cloud environments, highlighting its effectiveness on ensuring semantic interoperability for distributed AI orchestration.

## 1. Introduction

The recent advancements in large-scale AI models have reshaped the way in which intelligent systems are developed and deployed. Traditionally, AI workloads were executed in centralized cloud data centers; however, they are increasingly deployed across multiple layers of distributed infrastructures, including edge devices, fog nodes, and cloud platforms (Pazmiño-Ortiz, 2026), (Tuli, 2023). Within this context, Edge-Cloud Continuum provides a unified computing paradigm that seamlessly integrates Internet of Things (IoT) devices, edge servers, and centralized cloud resources into a single, continuous computational ecosystem (Ali, 2025). This architectural transition has been driven by the rapid growth of IoT devices and data generation, which makes transmitting and processing all data in centralized cloud infrastructures increasingly inefficient. At the same time, the applications with strong latency, reliability, and data privacy requirements have reinforced the need for distributed computing architectures (Escamilla-Ambrosio, 2018). Paradigms such as Federated Learning and on-device inference reflect this transition, enabling AI models to operate closer to data sources while preserving data locality (Zhan, 2025). At the same time, advances in model optimization and discretization techniques have made it feasible to deploy increasingly capable models at the edge, under strict computational and energy constraints. These developments have created a highly heterogeneous execution landscape, in which AI processes such as training, fine-tuning, or inference are distributed across edge, fog, and cloud layers (Ahmad, 2023), (Hasan, 2025). However, this distribution introduces significant challenges related to deployment, coordination, and interoperability. AI processes must operate across heterogeneous hardware platforms, execution environments, and orchestration frameworks, each with different assumptions about computational resources and data access.

*Corresponding author. E-mail address: tudor.cioara @cs.utcluj.ro, (T.Cioara).

Existing semantic models and ontologies provide only limited support for representing AI processes and their computational, distribution, and lifecycle features in an interoperable manner. Although widely adopted ontologies such as SAREF (García-Castro, 2023), (SAREF, 2025) have evolved through numerous domain-specific extensions, they remain focused on IoT entities and interactions, without providing semantic support for distributed AI workflows. SAREF has evolved into a comprehensive ontology suite comprising multiple domain-specific extensions and horizontal modules, such as SAREF4SYST (Lefrançois, 2023), (SAREF4SYS, 2020), representing interconnected systems and distributed topologies. It is also aligned with the W3C SSN/SOSA ontology (Janowicz, 2019) to support interoperability with semantic sensor models. However, these extensions remain focused on IoT systems and do not address the representation of AI workflows, their lifecycle, or their deployment across edge-fog-cloud environments. Ontologies in the field of AI, such as Artificial Intelligence Ontology (AIO) (Joachimiak, 2024), aim to standardize concepts, methodologies, and relationships between AI techniques, providing a coherent semantic framework for organizing and structuring knowledge in this field. However, they are limited to the conceptual level and do not model the execution of AI processes in distributed systems. Similarly, other recent approaches, such as the ontology proposed by Zhong et al. (Zhong, 2021) for federated learning systems and EdgeOnto proposed by Ghrab et al. (Ghrab, 2023), focus either on resource modeling and decision mechanisms or on IoT service orchestration, without providing a unified semantic framework that integrates AI processes, data flows, and distributed infrastructure. Therefore, existing ontological models cannot describe such distributed execution semantics, including dependencies between AI process stages, data flow constraints, and resource-aware deployment. Moreover, federated learning and on-device inference introduce new coordination and privacy-preserving requirements, where model updates are exchanged across nodes; therefore, semantic models are needed to represent these decentralized AI learning and execution patterns (Papadopoulos, 2024). Also, they lack support for AI process orchestration. This includes the specification of workflow structures, task dependencies, sequencing constraints, and conditional execution paths that are essential for orchestrating the AI pipelines (Kreuzberger, 2023). This leads to tightly coupled, platform-specific AI orchestration solutions, limiting interoperability and preventing automated reasoning over AI workflows (Eken, 2025).

Semantic interoperability presents an additional challenge in distributed AI environments. Computing nodes, AI processes, data assets, and communication flows are frequently described using heterogeneous metadata schemas and platform-specific representations, limiting unified interpretation and automated orchestration across the edge-fog-cloud continuum (Iftikhar, 2023). Existing semantic models and ontologies have significantly improved interoperability in cloud computing, edge computing, and IoT systems by providing standardized representations of infrastructure resources, services, devices, and data. However, these models primarily focus on infrastructure and service-level abstractions and provide only limited support for representing AI processes together with their computational, distribution, and lifecycle characteristics. Among the existing approaches, the Edge Cloud Computing Ontology (ECO) is particularly relevant because it models devices and computing resources spanning edge, fog, and cloud environments (Bernabé, 2022). Nevertheless, ECO does not explicitly conceptualize AI processes or their execution lifecycle, limiting its ability to support distributed AI workflows. Therefore, semantically equivalent AI resources and processes may still be represented inconsistently across platforms, hindering their discovery, composition, orchestration, and reuse.

In this respect, SAREF provides a relevant foundation for semantic interoperability by offering a common reference ontology for representing devices, functions, services, and their relationships, independently of the underlying communication technologies (García-Castro, 2023), (SAREF, 2025). SAREF supports the harmonization of heterogeneous representations and facilitates interoperable data exchange, integration, and reasoning across distributed systems by acting as a common semantic layer. These characteristics make SAREF a suitable core for extending semantic interoperability toward distributed AI workflows. However, AI processes introduce requirements that are not addressed by existing SAREF modules, including the representation of lifecycle stages such as training, validation, deployment, and inference, as well as the deployment and execution of AI workflows across heterogeneous edge, fog, and cloud resources (Pazmiño,

*Corresponding author. E-mail address: tudor.cioara @cs.utcluj.ro, (T.Cioara).

2026). The need to evolve SAREF to support emerging paradigms such as edge AI, federated learning, and generative AI has been recognized at the policy level in the European Commission ICT Rolling Plan (EU, 2026). Such an extension is needed because AI processes introduce specific requirements that go beyond traditional system modeling. These include the need to represent distinct lifecycle stages such as training, validation, deployment, inference, as well as the relationships between them (Steidl, 2023). Also, AI processes are inherently distributed; therefore, different activities of a single workflow may execute across heterogeneous environments, from resource-constrained edge devices to scalable cloud infrastructures (Pazmiño, 2026). Addressing this gap would enable a unified and machine-interpretable representation of AI workflows and resources across the edge-cloud continuum, supporting interoperability and facilitating automated reasoning, orchestration, and reuse of distributed AI systems.

To address the above limitations, this paper proposes DAI-ECC, a SAREF-compliant ontology for modeling distributed AI workflows across the edge-fog-cloud continuum. DAI-ECC extends the SAREF4SYST ontology, with concepts for representing AI pipelines, executable AI jobs, computational resources, deployment constraints, and communication relationships within heterogeneous computing environments. The ontology enables semantic interoperability, automated reasoning, and resource-aware orchestration of distributed AI workflows by defining a unified semantic model for both AI workflows and the edge-fog-cloud computational infrastructure. The ontology is formally specified based on the ETSI SAREF ontology engineering methodology (ETSI, 2024) and implemented using the SAREF-Pipeline (Lefrançois, 2023) ensuring consistency with existing SAREF design principles, reuse of established concepts, and interoperability within the SAREF ecosystem. The proposed ontology is evaluated through competency-question-based validation, SPARQL queries, and proof-of-concept scenarios involving smart grid applications and distributed AI workflows, demonstrating its ability to support workflow deployment, execution reasoning, and workload adaptation across heterogeneous edge, fog, and cloud infrastructures.

The main contributions of this paper are:

- A unified semantic model that integrates descriptions of AI workflows and heterogeneous computing infrastructures, enabling semantic interoperability and machine-interpretable resource descriptions.
- A SAREF-compliant ontology, extending the SAREF4SYST, with semantic concepts and relationships for distributed AI environments to explicitly represent interactions between infrastructure components, AI processes, and exchanged data within distributed systems, while benefiting from the interoperability and integration capabilities of the SAREF ecosystem.
- Demonstration of the ontology's applicability and expressiveness by modeling a federated learning scenario in smart energy grid environments and showcasing its role in supporting interoperability within an orchestration engine for managing AI processes.

The paper is structured as follows: Section 2 presents related work on ontology models for computational infrastructures and AI/ML artifacts. Section 3 introduces our methodology for ontology development, and Section 4 presents the DAI-ECC ontology, focusing on concepts for modeling distributed computational resources and AI processes. Section 5 reports the structural and proof-of-concept evaluation results based on a smart grid use case, while Section 6 concludes the paper and presents future work.

## 2. Related Work

Several ontologies have been proposed to support semantic interoperability in cloud computing, IoT, and edge computing by modeling computing resources, devices, services, and distributed infrastructures. Cloud computing ontologies aim at improving interoperability through the semantic description of cloud resources and services. CloudFNF (Al-Sayed, 2020) models cloud services by defining their functional and non-functional characteristics, supporting service description, discovery, and recommendation. Similarly, CoCoOn (Zhang, 2012) provides a semantic representation of cloud infrastructure services by modeling computing resources, storage, networking, and quality-of-service (QoS) attributes, enabling standardized service description and selection across providers. While these ontologies facilitate interoperability in cloud environments, they primarily focus on infrastructure resources and service provisioning, without explicitly representing AI processes or distributed AI workflows. Within IoT and edge computing ecosystems, ontologies have primarily been developed to support semantic interoperability among heterogeneous devices, sensors, services, and contextual information. IoT-Lite (Bermudez-Edo, 2016) introduces a lightweight semantic model for describing IoT devices, sensors, and observations, facilitating efficient data discovery across heterogeneous platforms. Similarly, FADA (Wu, 2022) proposes cloud-fog-edge architecture accompanied by an ontology that standardizes industrial data acquisition by modeling industrial equipment, communication protocols, and operational parameters. More generally, semantic models in this domain enable context-aware adaptation and reasoning over IoT infrastructures (Ryabinin, 2020). Among the most widely adopted semantic models for IoT interoperability is SAREF (García-Castro, 2023), which provides standardized abstractions for devices, services, and their relationships while remaining independent of underlying communication protocols. SAREF has evolved into a comprehensive ontology suite through numerous domain-specific extensions, including SAREF4ENER for smart energy (Daniele, 2016), SAREF4BLDG for buildings (Villalón, 2017), and SAREF4INMA for industrial manufacturing (de Roode, 2020). In addition, the horizontal module SAREF4SYST (Lefrançois, 2023) supports the representation of systems, their interconnections, and distributed topologies. SAREF is also aligned with the W3C SSN/SOSA ontology (Janowicz, 2019), providing interoperability with widely adopted semantic models for sensors and actuators. The Edge Cloud Computing Ontology (ECO) (Bernabé, 2022) is the closest to the present work. ECO models IoT systems composed of devices and data centers distributed across edge, fog, and cloud environments, introducing concepts such as computing nodes, middleware, software, services, and applications. The ontology builds upon the Smart Energy Aware Systems (SEAS) ontology, which models physical systems and their interrelationships through explicit representations of device connections. Since SEAS also served as a conceptual foundation for the development of SAREF4SYST, ECO defines a semantic basis for representing the edge-cloud continuum. However, its primary focus remains the description of computing infrastructures and deployed applications. Although these ontologies offer comprehensive support for modeling IoT ecosystems, they do not explicitly represent AI workflows, their execution lifecycle, or the distribution of AI processes across heterogeneous computing environments.

*Corresponding author. E-mail address: tudor.cioara @cs.utcluj.ro, (T.Cioara).

In parallel with ontologies for distributed computing infrastructures, several semantic models have been proposed to represent AI concepts, machine learning processes, models, experiments, and evaluation results. Existing efforts aim to facilitate knowledge sharing, reproducibility, interoperability, and model management. Examples include ML-Schema (Publio, 2018), XMLPO (Xhani, 2024), the ontology for explainable ML workflows (Nakagawa, 2021), MLOnto (Braga, 2020), ITO (Blagec, 2022), VIS4ML (Sacha, 2019), and FIDES (Fernandez, 2023), which provide semantic representations of different aspects of the AI/ML domain. ML-Schema (Publio, 2018) provides a top-level semantic model for representing metadata about machine learning processes, including datasets, features, algorithms, implementations, hyperparameters, models, evaluations, runs, and experiments. By offering a shared semantic structure, it supports interoperability across ML platforms. In the same direction, the ontology for the semantic description of explainable ML workflows (Nakagawa, 2021) models ML workflows, operations, algorithms, parameter settings, input/output data, ML models, model evaluations, generated explanations, and explainability evaluations. Building on work XMLPO (Xhani, 2024) extends Explainable ML Workflows ontology and reuse ML-Schema to model explainable machine learning pipelines. It represents the main stages of an ML pipeline, such as data input, data preparation, preprocessing, model training and testing, and model evaluation, as well as XAI-related elements, including explanation approaches, explanation formats, explanation scope, and evaluation metrics. MLOnto (Braga, 2020) represents general knowledge about the ML domain by organizing ML concepts into high-level classes such as algorithms, applications, dependencies, frameworks, dictionaries, involved actors, and ML types, including supervised, unsupervised, semi-supervised, reinforcement learning, and AutoML. The Intelligence Task Ontology and

*Corresponding author. E-mail address: tudor.cioara @cs.utcluj.ro, (T.Cioara).

Knowledge Graph (ITO) (Blagec, 2022) represent AI tasks, benchmark datasets, AI models, and performance metrics, and relates AI tasks to input/output data, benchmark datasets, software models, topics, and quantitative performance measures. VIS4ML (Sacha, 2019) represents visual analytics-assisted machine learning workflows through processes and input/output entities, including data, models, and human knowledge. It captures how visual analytics supports activities such as data preparation, model understanding, feature and parameter analysis, learning process monitoring, result analysis, and model comparison. FIDES (Fernandez, 2023) focus on the accountability of machine learning systems by modelling estimations, results, and contextual information through the AffectedBy, Execution-Executor-Procedure, and Result-Context ontology design patterns. It also relies on ML-Schema to represent ML-specific elements such as datasets, features, algorithms, implementations, hyperparameters, models, and evaluations.

Overall, existing ontologies address complementary aspects of distributed computing, including cloud services, IoT devices, AI artefacts, and edge-cloud infrastructures. However, none provides an integrated semantic model capable of representing AI workflows together with their computational requirements, lifecycle, communication patterns, and deployment across the edge-fog-cloud continuum. To address this gap, we propose DAI-ECC, which extends the SAREF ecosystem to support the semantic representation of distributed AI workflows. The proposed ontology provides a unified semantic model that enables interoperability, automated reasoning, and resource-aware orchestration of AI workflows across heterogeneous edge, fog, and cloud environments.

## 3. Ontology development methodology

For the development of the DAI-ECC ontology, we followed an adapted version of the Linked Open Terms (LOT) methodology, which consists of four main steps: (i) ontology requirements specification, (ii) ontology implementation, (iii) ontology publication, and (iv) ontology maintenance. Figure 1 provides an overview of the ontology development methodology.

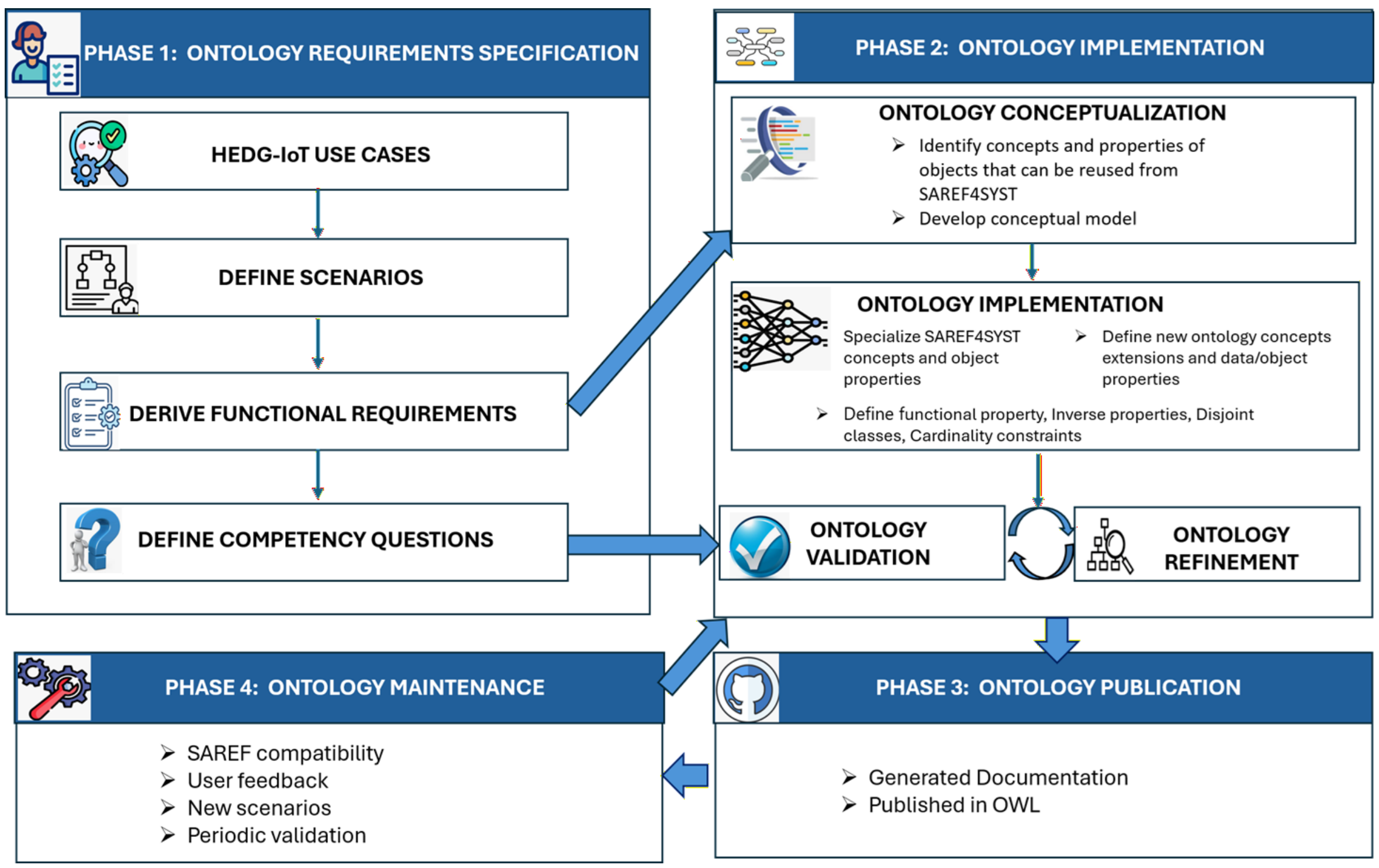

Figure 1. DAI-ECC ontology development methodology

*Corresponding author. E-mail address: tudor.cioara @cs.utcluj.ro, (T.Cioara).

The *first phase of the methodology, Ontology Requirements Specification*, aims to define the scope, objectives, intended users, and functional requirements of the ontology before its formal implementation. The requirements specification of DAI-ECC was driven by the use cases developed within the HEDG-IoT project (HEDGE-IoT, 2026), which develops distributed AI solutions for smart grid applications deployed across the edge-fog-cloud continuum. In this context, we have designed an orchestrator for AI-driven energy services in the smart grid, exposing the need for interoperable management of heterogeneous computational infrastructures and AI workflows. First, computational resources distributed across edge, fog, and cloud environments are described using heterogeneous representations, which limit resource discovery and management. Second, AI workflows, including their execution stages, resource requirements, and communication patterns, lack a common semantic representation, making automated deployment, orchestration, and adaptation across heterogeneous infrastructures difficult. To capture these requirements systematically, three representative application scenarios were identified (see Table 1).

Table 1. Use case and scenarios for ontology requirements specification

| ***Narrative of Use Case*** |
|---|
| ***Short description*** |
| The use case focuses on enabling the automated coordination, management, and execution of AI-enabled tasks through an orchestrator. The orchestrator leverages swarm-based algorithms to discover, select, and optimize resource usage, ensuring both computational and communication efficiency. It manages deployment, coordination, and resource allocation and supports federated hyperparameter tuning and training optimization. Additionally, it enables services to roll out at the edge, allowing automated updates and deployment of new versions. Integration with the Eclipse Data Space Connector (Eclipse Foundation, 2021) ensures compliance with data space standards and secure, interoperable data and service exchange. |
| ***Complete description*** |
| **The use case was described using three scenarios:**<br>**Scenario 1: Energy services orchestration at the edge for responsiveness and geographic redundancy.** Containerized energy services are deployed across edge-fog-cloud distributed infrastructure overlapping the smart grid. The services and the infrastructure available computing nodes are registered with the computational orchestrator. The orchestrator continuously monitors service locations, resource availability, and task assignments. An integrated Kubernetes component handles the initial task allocation based on current resource availability and predefined configurations. It also supports live monitoring of service status and system resources. When predefined events occur, such as violations of service level agreements or policy conditions, the orchestrator responds by executing a swarm-based optimization algorithm to determine which services should be migrated to other nodes. During service migration, the persistent state and data of each service must also be transferred, and their connectivity via the Eclipse Data Space Connector must be maintained to ensure secure and interoperable data exchange. Therefore, it ensures service responsiveness and geographic redundancy.<br>**Scenario 2: Federated AI-driven energy services orchestration for cost-effective data exchanges.** The orchestrator manages federated learning processes initiated by AI services deployed across edge and fog/cloud nodes. The federated architecture may follow either a hierarchical or peer-to-peer model. All participating nodes are registered with the orchestrator, providing metadata on their computational capabilities and availability. Based on service-specific requirements, the orchestrator can cluster nodes to enhance training efficiency. It also performs hyperparameter tuning and training optimization using heuristic-based algorithms, ensuring efficient use of distributed resources at the edge and minimizing data exchange overhead among edge and fog/cloud nodes.<br>**Scenario 3: Energy service rolling out at the edge.** The orchestrator can support service providers, such as system operators, to automatically roll out new versions of energy services for the consumers. The data models (packages or files) are sent through the Eclipse Data Space Connector. It detects available updates for application components, manages versioning, and handles service interruption during updates. |

Each scenario was analyzed to identify the entities involved, their relationships, the information exchanged among them, and the reasoning capabilities required to support orchestration decisions. This analysis led to the identification of the core ontology concepts, including computational nodes, AI pipelines, AI jobs, execution resources, deployment constraints, communication interfaces, and workflow dependencies.

*Corresponding author. E-mail address: tudor.cioara @cs.utcluj.ro, (T.Cioara).

Additionally, a set of functional requirements was derived to define the capabilities expected from the ontology. These requirements specify the information that must be represented to support semantic interoperability across distributed AI infrastructures. In particular, the ontology must enable the representation of heterogeneous computing nodes and their capabilities, AI workflows and their constituent tasks, execution requirements and deployment constraints, data exchanges between workflow stages, and the relationships between AI processes and the computational resources on which they execute. The functional requirements subsequently guided the formulation of competency questions, which provide a formal mechanism for validating that the ontology satisfies its intended purpose (see Table 2). Each competency question expresses a query that the ontology should be capable of answering in the context of the smart grid scenarios. The competency questions therefore establish measurable validation criteria and ensure that the resulting ontology supports the orchestration, deployment, migration, and monitoring of distributed AI workflows across heterogeneous edge, fog, and cloud environments.

Table 2. CQs derived for ontology validation

| **Scope** | **Competency Question** | **Expected Answer** |
|---|---|---|
| AI Workflow Lifecycle | Which AIJobs belong to a given AIPipeline? | Aggregation_A, Validation_A, Preprocess_A, Preprocess_B, … |
| | What data object does each stage produce, and of what type? | Aggregation_A→ GlobalModel (Model), LocalTrainingB → ModelUpdateX(ModelUpdate),… |
| | What is the execution order of the pipeline, and which stage is a conditional branch point? | Preprocess → Train → Aggregate → Validate → Inference \| Retrain |
| AI data governance | Which data is exposed across the organisational boundary through a data-space connector? | The local model updates A, B, C, … |
| | Are there any raw datasets exposed across the boundary? | No, all raw datasets stay local |
| | For the federated aggregation, which data is consumed, from which node does it originate, and at what latency? | Model Update A from node X with latency 5.2 ms … |
| Computational Infrastructure | Which deployment locations satisfy both the latency and computational constraints of an AIJob? | Node X, Node Y, Node Z, … |
| | Can every AIJob in an AIPipeline be deployed on the available infrastructure? | AIJob A → Node X, AIJob B → Node Y, … |
| | Which pipelines require reconfiguration due to computational infrastructure changes? | Pipeline_A → AIJob AX, AY, AZ, Pipeline_B → AIJob BX, BY, BZ, … |

*The second phase, Ontology Implementation,* transformed the requirements identified in the previous step into a formal semantic model. DAI-ECC was implemented as an extension of the ETSI SAREF4SYST ontology, reusing its generic concepts for representing systems, connections, and connection points as the foundation for modeling edge-fog-cloud computational infrastructures. The implementation process consisted of four main activities: (i) conceptual modeling, (ii) ontology formalization, (iii) ontology refinement, and (iv) validation. Initially, a conceptual model was developed using draw.io to identify the ontology classes, object properties, data properties, and their relationships based on the functional requirements and competency questions derived during the requirements specification phase. The conceptual model was then formalized in OWL using Protégé, where the ontology classes, properties, restrictions, and axioms were implemented while extending the concepts provided by SAREF4SYST. Following the formalization, the ontology was refined through iterative revisions to ensure that competency questions could be answered using the conceptual model. Reasoning capabilities were incorporated to support inference over distributed AI workflows and computational infrastructures, enabling the verification of scenarios such as federated AI pipeline execution, resource-aware deployment, and workload migration across heterogeneous edge, fog, and cloud resources. Finally, the ontology implementation was validated by assessing its ability to answer the competency questions defined during the requirements specification phase (see Section 5). Each competency question was translated into a SPARQL query and executed over representative smart grid scenarios to verify that the required information could be retrieved correctly.

*Corresponding author. E-mail address: tudor.cioara @cs.utcluj.ro, (T.Cioara).

*In the third phase, ontology publication,* we have provided it in OWL using persistent namespaces and followed the naming conventions and engineering guidelines of the ETSI SAREF ecosystem. ETSI SAREF-Pipeline automatically generates the ontology documentation, diagrams, and documentation from the ontology sources, facilitating reuse, inspection, and future extension. The ontology, its documentation, and illustrative examples are made publicly available through a version-controlled repository to promote transparency, reproducibility, and community reuse.

*The fourth phase, Ontology Maintenance,* ensures that DAI-ECC can evolve as distributed AI technologies and edge-fog-cloud infrastructures continue to develop. Since the ontology extends SAREF4SYST, future updates will preserve compatibility with new releases of the SAREF ecosystem while incorporating additional concepts required by federate learning, generative AI, or edge intelligence. Maintenance activities include adaptation to user feedback, refinement of ontology concepts, addition of scenarios, and periodic validation of logical consistency and interoperability with other semantic models.

## 4. DAI-ECC ontology

Figure 2 presents the conceptual model of DAI-ECC, showing the core concepts reused from SAREF4SYST, the extensions introduced to represent edge-fog-cloud infrastructures, and the newly defined concepts for modelling distributed AI workflows. The reused SAREF4SYST concepts are highlighted in pink, the infrastructure-related extensions are shown in blue, and the AI workflow-specific concepts are highlighted in yellow. The figure also shows the object properties that capture how infrastructure components, computational resources, data objects, workloads, and AI processes interact within distributed environments.

The core concept of SAREF4SYST is *s4syst:System*, which represents a system conceptually separated from its environment, but able to interact with other systems. Systems can be organized hierarchically through the transitive properties *s4syst:hasSubSystem* and *s4syst:subSystemOf*, which support the modelling of system decomposition and aggregation. Interactions between systems are represented through class *s4syst:Connection*. *Systems* are linked to *Connections* using *s4syst:connectedThrough*, while *Connections* are linked to the *Systems* through *s4syst:connectsSystem*. For simpler assertions, *Systems* may also be directly related through the symmetric property *s4syst:connectedTo*. SAREF4SYST also defines *s4syst:ConnectionPoint*, which represents the specific point at which a system participates in a connection. Each *ConnectionPoint* is associated with exactly one *System* through *s4syst:connectionPointOf*. A *Connection* is then linked to the relevant *ConnectionPoints* through *s4syst:connectsSystemAt* and *s4syst:connectsSystemThrough*. In DAI-ECC, these SAREF4SYST concepts and relations are specialized in representing the distributed computational nodes, network links, interfaces, data connection points, and other infrastructure elements required for distributed AI processes.

*Corresponding author. E-mail address: tudor.cioara @cs.utcluj.ro, (T.Cioara).

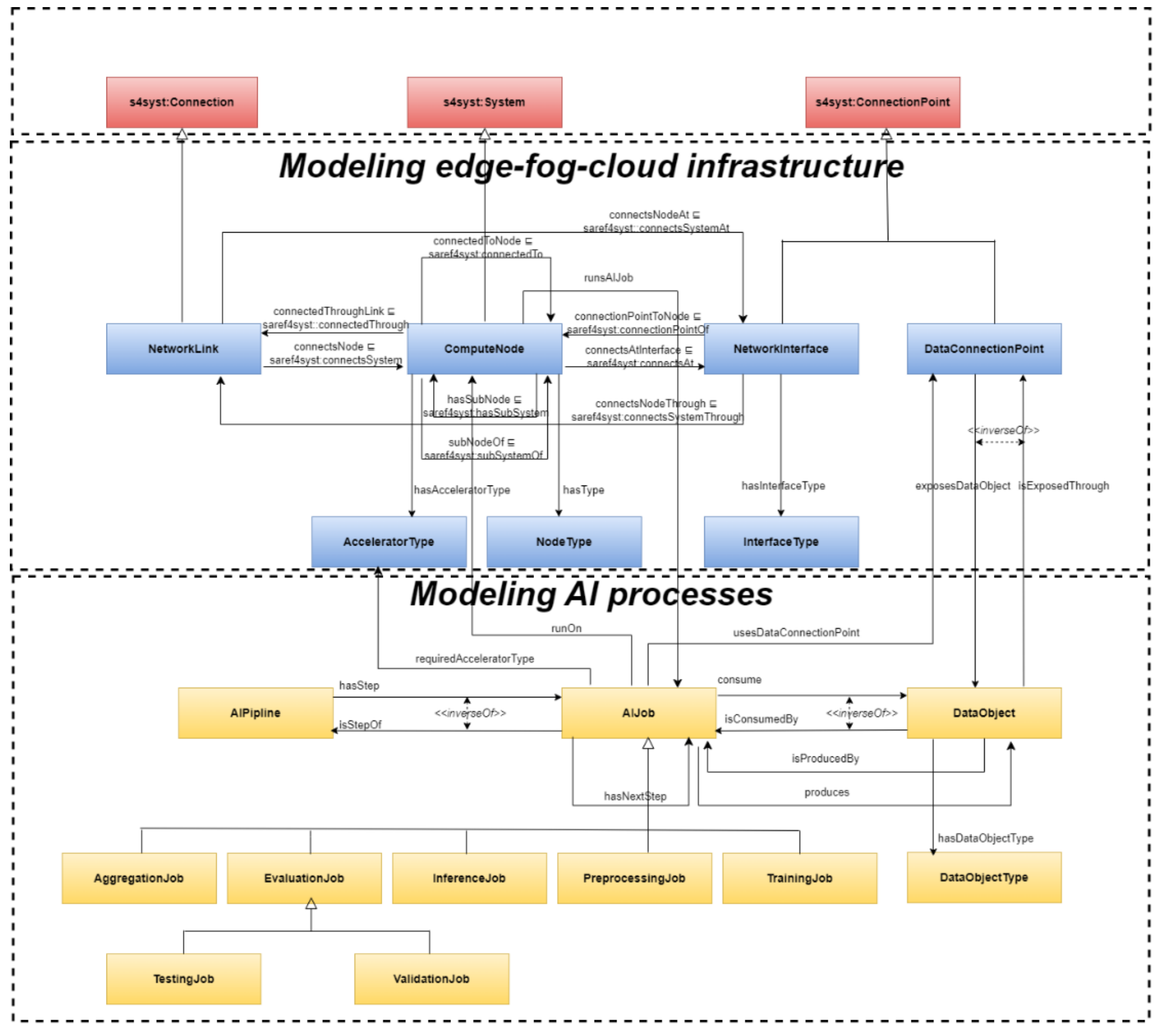

Figure 2. Conceptual model of DAI-ECC ontology

## 4.1. Modeling Distributed Infrastructures

The main concept used to model a compute node in ontology is *ComputeNode*, which is defined as a specialization of s4syst:*System*. A *ComputeNode* represents a computational entity on which distributed AI processes are executed. To represent the heterogeneity of the computation infrastructure, DAI-ECC introduces the *NodeType* concept, which classifies computing nodes according to the layer in which they operate. In the current ontology, this classification is instantiated through *Edge*, *Fog*, and *Cloud* individuals. This distinction is important because, when an AI Job is placed for execution on a compute node, the placement decision should consider data proximity, latency constraints, processing capacity, and the role of the node within the distributed execution pipeline. In addition to this classification, each *ComputeNode* is described through data properties that model the computational resources of the node. These properties describe the computing node in terms of *processing capacity*, *memory*, *storage*, *accelerator* features, and *location*. They enable the requirements of AI jobs and workloads to be compared with the capabilities of available compute nodes, thereby supporting resource aware deployment and workload migration decisions. The data properties associated with a *Compute Node* are summarized in Table 3.

*Corresponding author. E-mail address: tudor.cioara @cs.utcluj.ro, (T.Cioara).

Table 3. Data properties used to describe *ComputeNode* concept

| Data property | DL axiom for domain and range | Description |
|---|---|---|
| hasCpuCores | ∃ hasCpuCores.xsd:integer ⊑ ComputeNode | # CPU cores available on the node |
| hasGpuCount | ∃ hasGpuCount.xsd:integer ⊑ ComputeNode | # GPUs available on the node |
| hasAcceleratorCount | ∃ hasAcceleratorCount.xsd:integer ⊑ ComputeNode | # Hardware accelerators available on the node |
| hasAcceleratorModel | ∃ hasAcceleratorModel.xsd:string ⊑ ComputeNode | Type or model of the accelerator |
| hasAcceleratorMemory | ∃ hasAcceleratorMemory.xsd:decimal ⊑ ComputeNode | Memory capacity of the accelerator |
| hasMemory | ∃ hasMemory.xsd:decimal ⊑ ComputeNode | Memory capacity of the node |
| hasStorage | ∃ hasStorage.xsd:decimal ⊑ ComputeNode | Storage capacity of the node |
| hasLocation | ∃ hasLocation.xsd:string ⊑ ComputeNode | Location of the node |

To represent interactions between compute nodes, DAI-ECC extends the SAREF4SYST with the *NetworkLink*, *NetworkInterface*, and *DataConnectionPoint* concepts. Together, these concepts provide the semantic basis for representing communication links, network interfaces, and data access points, between *Compute Nodes* within distributed *AI pipelines*. The *NetworkLink* is defined as a specialization of s4syst:*Connection* concept and represents communication links between *ComputeNode* participating in the execution of distributed AI processes. Since communication conditions can directly affect distributed AI Job execution, each *NetworkLink* is described through data properties (see Table 4) that describe its communication profile, such as *bandwidth*, *throughput*, *latency*, *jitter*, and *packet loss rate*.

Table 4. Data properties used to describe *NetworkLink* concept

| Data property | DL axiom | Description |
|---|---|---|
| hasBandwidth | ∃ hasBandwidth.xsd:decimal ⊑ NetworkLink | Nominal data transfer capacity of the network link |
| hasThroughput | ∃ hasThroughput.xsd:decimal ⊑ NetworkLink | Effective rate of data transmission over network link |
| hasLatency | ∃ hasLatency.xsd:decimal ⊑ NetworkLink | Transmission delay associated with the network link |
| hasJitter | ∃ hasJitter.xsd:decimal ⊑ NetworkLink | Variation in packet transmission delay |
| hasPacketLossRate | ∃ hasPacketLossRate.xsd:integer ⊑ NetworkLink | Proportion of packets lost during transmission |

The access of computing nodes to the communication infrastructure is represented through the *NetworkInterface* concept, defined as a specialization of s4syst:*ConnectionPoint*. A *NetworkInterface* represents the communication access point through which a *ComputeNode* connects to a *NetworkLink* and exchanges data with other nodes in the distributed infrastructure. Each *NetworkInterface* is described through data properties, summarized in Table 5, that describe the identification and communication characteristics, such as *IP address*, *MAC address*, *maximum supported speed*, and m*aximum transmission unit* (MTU). These properties allow the ontology to represent not only that a compute node is connected to the communication infrastructure, but also the technical characteristics of the interface through which communication is performed. This information is relevant for distributed AI pipelines because interface characteristics can affect the transfer of data objects, model updates, global models, and intermediate results between pipeline stages. Each *NetworkInterface* is associated with a single *ComputeNode*, reflecting the fact that a communication interface belongs to a specific computational entity.

*Corresponding author. E-mail address: tudor.cioara @cs.utcluj.ro, (T.Cioara).

Table 5. Data properties used to decribe *NetworkInterface* concept

| Data property | DL axiom | Description |
|---|---|---|
| hasIPAddress | ∃ hasIPAddress.xsd:string ⊑ NetworkInterface | IP address associated with the network interface |
| hasMacAddress | ∃ hasMacAddress.xsd:string ⊑ NetworkInterface | MAC address associated with the network interface |
| hasMaxSpeed | ∃ hasMaxSpeed.xsd:decimal ⊑ NetworkInterface | Maximum transfer speed supported by the interface |
| hasMTU | ∃ hasMTU.xsd:integer ⊑ NetworkInterface | Maximum transmission unit supported by the interface |

To describe the communication technology used by a *NetworkInterface*, DAI-ECC introduces the *InterfaceType* concept. This classifies network interfaces into types such as Ethernet, WiFi, Fiber, 4G, 5G, and LoRaWAN. It adds contextual information to *NetworkInterface* data properties and supports the modelling of communication conditions that may influence data transfer, model update exchange, and the coordination of AI jobs across distributed infrastructures. The distinction is relevant because different interface types imply different communication capabilities. Wired technologies, such as Ethernet and Fiber, usually support stable or high-capacity communication between fixed edge, fog, and cloud nodes, while wireless technologies, such as WiFi, 4G, 5G, and LoRaWAN, support mobile, wireless, or long-range communication.

Data accessibility within the distributed infrastructure is represented through the *DataConnectionPoint* concept, defined as a specialization of *s4syst:ConnectionPoint*. A *DataConnectionPoint* models the access point through which data objects are exposed, consumed, or exchanged by AI processes. This concept enables ontology to distinguish between communication-level connectivity and data-level accessibility, which is particularly relevant for distributed AI pipelines relying on the exchange of datasets, model updates, global models, or intermediate results. To support interoperability with energy dataspace infrastructures, DAI-ECC defines concepts like *EclipseDataSpaceConnectionPoint*. This concept represents data access points managed through Eclipse Dataspace mechanisms, enabling the ontology to model data exchange settings in which access to AI-related data is governed by dataspace connectors, policies, or controlled sharing mechanisms.

DAI-ECC also introduces the *AcceleratorType* concept to classify the hardware accelerators available on a *ComputeNode*. *AcceleratorType* is represented through individuals such as GPU, TPU, NPU, and FPGA. Modelling accelerator types explicitly is important because AI jobs may depend on specific hardware capabilities, such as GPU-based parallel processing, NPU-based edge inference, or FPGA-based low-latency execution. Therefore, *AcceleratorType* helps identify compute nodes whose hardware capabilities match the processing and memory requirements of a given AI job.

Beyond the class hierarchy and data properties, DAI-ECC defines a set of object properties that capture relationships within the distributed infrastructure. These properties connect compute nodes with network links, network interfaces, data access points, node types, interface types, and accelerator types. They also allow the modelling of the direct connectivity between nodes and the hierarchical composition of distributed infrastructures. At the infrastructure level, object properties such as *connectedThroughLink*, *connectsNode*, and *connectedToNode* describe how *ComputeNode* instances are connected through *NetworkLink* instances. The relationships at the interface level are represented through properties such as *connectsAtInterface*, *connectionPointToNode*, *connectsNodeAt*, and *connectsNodeThrough*, which associate compute nodes, network interfaces, and communication links. In addition, properties such as *hasSubNode* and *subNodeOf* model hierarchical infrastructures of Compute Nodes, while *hasType*, *hasInterfaceType*, and *hasAcceleratorType* specify the type of a compute node, the type of network interface, and the type of hardware accelerator available on a compute node. The main object properties used for infrastructure modelling are summarized in Table 6 together with their corresponding DL axioms and formal characteristics.

*Corresponding author. E-mail address: tudor.cioara @cs.utcluj.ro, (T.Cioara).

Table 6. Object properties used for modeling the computing infrastructure and their corresponding DL axioms

| Object property | Axioms for domain and range | Characteristics |
| --- | --- | --- |
| connectedThroughLink ⊑ saref4syst:connectsThrough | ∃ connectedThroughLink.⊤ ⊑ ComputeNode<br>⊤ ⊑ ∀ connectedThroughLink.NetworkLink | N/A |
| connectedToNode ⊑ saref4syst:connectedTo | ∃ connectedToNode.⊤ ⊑ ComputeNode<br>⊤ ⊑ ∀ connectedToNode.ComputeNode | connectedToNode ≡ connectedToNode$^{-}$ |
| connectionPointToNode ⊑ connectionPointTo | ∃ connectionPointToNode.⊤ ⊑ NetworkInterface<br>⊤ ⊑ ∀ connectionPointToNode.ComputeNode | connectionPointToNode ≡ connectsAtInterface$^{-}$<br>⊤ ⊑ ≤ 1 connectionPointToNode |
| connectsAtInterface ⊑ saref4syst:connectsAt | ∃ connectsAtInterface.⊤ ⊑ ComputeNode<br>⊤ ⊑ ∀ connectsAtInterface.NetworkInterface | connectsAtInterface ≡ connectionPointToNode$^{-}$<br>⊤ ⊑ ≤ 1 connectsAtInterface$^{-}$ |
| connectsNode ⊑ saref4syst:connectsSystem | ∃ connectsNode.⊤ ⊑ NetworkLink ⊤ ⊑ ∀ connectsNode.ComputeNode | N/A |
| connectsNodeAt ⊑ saref4syst:connectsSystemAt | ∃ connectsNodeAt.⊤ ⊑ NetworkLink<br>⊤ ⊑ ∀ connectsNodeAt.NetworkInterface | N/A |
| connectsNodeThrough ⊑ saref4syst:connectsSystemThrough | ∃ connectsNodeThrough.⊤ ⊑ NetworkInterface<br>⊤ ⊑ ∀ connectsNodeThrough.NetworkLink | N/A |
| hasSubNode ⊑ saref4syst:hasSubSystem | ∃ hasSubNode.⊤ ⊑ ComputeNode<br>⊤ ⊑ ∀ hasSubNode.ComputeNode | hasSubNode ≡ subNodeOf$^{-}$ |
| subNodeOf ⊑ saref4syst:subSystemOf | ∃ subNodeOf.⊤ ⊑ ComputeNode<br>⊤ ⊑ ∀ subNodeOf.ComputeNode | subNodeOf ≡ hasSubNode$^{-}$ |
| hasType | ∃ hasType.⊤ ⊑ ComputeNode<br>⊤ ⊑ ∀ hasType.NodeType | ⊤ ⊑ ≤ 1 hasType |
| hasInterfaceType | ∃ hasInterfaceType.⊤ ⊑ NetworkInterface<br>⊤ ⊑ ∀ hasInterfaceType.InterfaceType | ⊤ ⊑ ≤ 1 hasInterfaceType |
| hasAcceleratorType | ∃ hasAcceleratorType.⊤ ⊑ ComputeNode<br>⊤ ⊑ ∀ hasAcceleratorType.AcceleratorType | N/A |

## 4.2. Modelling AI Processes

DAI-ECC models distributed AI processes through two core concepts: *AIPipeline* and *AIJob*. An *AIPipeline* represents an AI workflow composed of multiple *AIJob*, while an *AIJob* represents a single executable AI task within a workflow. The execution requirements and deployment constraints of an *AIJob*, including maximum allowed latency, required execution location, required CPU cores, required memory, minimum bandwidth, required accelerators, and accelerator memory, are described through data properties presented in Table 7. These properties allow for assessing whether an *AIJob* can be deployed on a specific *ComputeNode* that satisfies its resource and execution constraints.

The *AIJob* concept is specialized into several types of jobs that represent the main phases of an AI pipeline: *PreprocessingJob*, *TrainingJob*, *InferenceJob*, *EvaluationJob*, and *AggregationJob*. These subclasses allow the ontology to model different execution scenarios, including centralized learning, federated learning, and distributed inference. The *EvaluationJob* class is further specialized into *ValidationJob* and *TestingJob*, enabling the ontology to distinguish between validation activities used during model development and testing activities used for final performance assessment. To avoid ambiguous modeling of pipeline steps, the main *AIJob* subclasses are declared mutually disjoint. This ensures that an individual job cannot be simultaneously interpreted, for example, as both a *TrainingJob* and an *InferenceJob*.

*Corresponding author. E-mail address: tudor.cioara @cs.utcluj.ro, (T.Cioara).

Similarly, *ValidationJob* and *TestingJob* are declared disjoint within *EvaluationJob*. These disjointness axioms support consistency checking and help detect modeling errors during ontology validation.

Table 7. Data properties of the *AIJob* concept

| Data property | Axiom for domain and range |
|---|---|
| maxAllowedLatency | ∃ maxAllowedLatency.xsd:decimal ⊑ AIJob |
| requiredLocation | ∃ requiredLocation.xsd:string ⊑ AIJob |
| requiredCpuCores | ∃ requiredCpuCores.xsd:integer ⊑ AIJob |
| requiredMinBandwidth | ∃ requiredMinBandwidth.xsd:decimal ⊑ AIJob |
| requiredRamGB | ∃ requiredRamGB.xsd:decimal ⊑ AIJob |
| requiredAcceleratorMemory | ∃ requiredAcceleratorMemory.xsd:decimal ⊑ AIJob |
| requiredAcceleratorCount | ∃ requiredAcceleratorCount.xsd: integer ⊑ AIJob |

To represent the data entities manipulated throughout a distributed AI pipeline, DAI-ECC introduces the *DataObject* concept. A *DataObject* denotes any data entity that may be consumed, produced, or exchanged by an *AIJob*, including datasets, trained models, model updates, and intermediate results. This concept allows the ontology to capture the data flow between pipeline stages. For example, a preprocessing job may consume a raw dataset and produce a cleaned dataset, a training job may consume the cleaned dataset and produce a trained model, while an aggregation job may consume model updates generated by distributed training nodes. To distinguish the role of each data entity within the AI workflow, DAI-ECC defines the *DataObjectType* concept, instantiated through *Dataset*, *Model*, and *ModelUpdate* individuals. This enables the ontology to differentiate between data used for preprocessing or training, models resulting from execution, and model updates exchanged in federated learning scenarios. In addition, access constraints associated with data objects are represented through the *accessPolicy* data property. This property allows the ontology to specify restrictions related to data use, sharing, or access, which is particularly relevant in distributed AI settings where data and models may move across organizational or computational boundaries.

To connect *AIJob* instances with the distributed infrastructure and to describe their execution within an *AI pipeline*, DAI-ECC defines a set of object properties, presented in Table 6. The execution of *AI jobs* on the distributed infrastructure is represented by the *runsAIJob* and *runsOn* properties. The *runsAIJob* property links a *ComputeNode* to an *AIJob*, indicating that a particular *Compute Node* is executing an *AI job.* The inverse relationship, *runsOn*, allows the representation of the node on which a particular job is executed. Access to data resources is modeled by the *usesDataConnectionPoint*, *exposesDataObject*, and *isExposedThrough* properties. The *usesDataConnectionPoint* property links an *AIJob* to the *DataConnectionPoint* used to obtain the resources required for execution. In addition, *exposesDataObject* associates a *DataConnectionPoint* with a *DataObject* available through that point, while the inverse relationship, *isExposedThrough*, indicates the *DataConnectionPoint* through which a *DataObject* can be accessed. The data flow between AI jobs and data objects is described by *consumesData*, *isConsumedBy*, *producesData*, and *isProducedBy* properties. The *consumesData* property indicates that an *AIJob* uses a particular *DataObject* as input, and the inverse relationship, *isConsumedBy*, links the *DataObject* to the *AIJob* that consumes it. Similarly, the *producesData* property links an *AIJob* to the *DataObject* generated as a result of its execution, while *isProducedBy* indicates the inverse relation, linking a *DataObject* to the *AIJob* that produced it. The classification of data objects is achieved through the *hasDataObjectType* property, which links each *DataObject* to a *DataObjectType*. The *DataObjectType* concept is instantiated through individuals such as *Dataset*, *Model*, and *ModelUpdate*, allowing the ontology to distinguish between input datasets, trained models, and model updates exchanged within an AI pipeline.

The relationship between an *AIJob* and an *AIPipeline* is modeled through the *hasStep*, *isStepOf*, and *hasNextStep* properties. The *hasStep* property links an *AIPipeline* to the *AIJob* instances that form its steps. The inverse property, *isStepOf*, links each *AIJob* back to the *AIPipeline* to which it belongs. The *hasNextStep* property represents the execution sequence by linking one *AIJob* to the job that follows it in the same pipeline. Together, these properties allow the ontology to describe both the composition of an AI pipeline and the order in which its jobs are executed. In addition to the properties that describe pipeline composition and execution order, DAI-ECC defines the *requiresAcceleratorType* property, which links an *AIJob* to the *AcceleratorType* required for its execution, such as GPU, TPU, NPU, or FPGA.

*Corresponding author. E-mail address: tudor.cioara @cs.utcluj.ro, (T.Cioara).

Table 8: Object properties used for modeling AIworklow and their corresponding DL axioms

| Object Property | Domain and Range | Characteristics / Axioms |
|---|---|---|
| runAIJob | ∃runAIJob.⊤ ⊑ ComputeNode ⊤ ⊑ ∀runAIJob.AIJob | runAIJob ≡ runOn⁻ |
| runOn | ∃runOn.⊤ ⊑ AIJob ⊤ ⊑ ∀runOn.ComputeNode | runOn ≡ runAIJob⁻ |
| exposesDataObject | ∃exposesDataObject.⊤ ⊑ DataConnectionPoint ⊤ ⊑ ∀exposesDataObject.DataObject | exposesDataObject ≡ isExposedThrough⁻ |
| isExposedThrough | ∃isExposedThrough.⊤ ⊑ DataObject ⊤ ⊑ ∀isExposedThrough.DataConnectionPoint | isExposedThrough ≡ exposesDataObject⁻ |
| usesDataConnectionPoint | ∃usesDataConnectionPoint.⊤ ⊑ AIJob ⊤ ⊑ ∀usesDataConnectionPoint.DataConnectionPoint | — |
| consumesData | ∃consumesData.⊤ ⊑ AIJob ⊤ ⊑ ∀consumesData.DataObject | consumesData ≡ isConsumedBy⁻ usesDataConnectionPoint ∘ exposesDataObject ⊑ consumesData |
| isConsumedBy | ∃isConsumedBy.⊤ ⊑ DataObject ⊤ ⊑ ∀isConsumedBy.AIJob | isConsumedBy ≡ consumesData⁻ |
| producesData | ∃producesData.⊤ ⊑ AIJob ⊤ ⊑ ∀producesData.DataObject | producesData ≡ isProducedBy⁻ |
| isProducedBy | ∃isProducedBy.⊤ ⊑ DataObject ⊤ ⊑ ∀isProducedBy.AIJob | isProducedBy ≡ producesData⁻ |
| hasDataObjectType | ∃hasDataObjectType.⊤ ⊑ DataObject ⊤ ⊑ ∀hasDataObjectType.DataObjectType | ⊤ ⊑ ≤1 hasDataObjectType.⊤ |
| hasNextStep | ∃hasNextStep.⊤ ⊑ AIJob ⊤ ⊑ ∀hasNextStep.AIJob | — |
| hasStep | ∃hasStep.⊤ ⊑ AIPipeline ⊤ ⊑ ∀hasStep.AIJob | hasStep ≡ isStepOf⁻ |
| isStepOf | ∃isStepOf.⊤ ⊑ AIJob ⊤ ⊑ ∀isStepOf.AIPipeline | isStepOf ≡ hasStep⁻ |
| requiredAcceleratorType | ∃ requiredAcceleratorType.⊤ ⊑ AIJob ⊤ ⊑ ∀ requiredAcceleratorType. AcceleratorType | |

## 5. Evaluation results

This section demonstrates the capabilities of DAI-ECC through representative smart grid scenarios. The ontology was integrated with the HEDGE-IoT energy services orchestrator to support the management of AI-driven workflows across heterogeneous edge–fog–cloud infrastructures. By providing a unified semantic representation of both the computational infrastructure and the AI workflow, DAI-ECC enables semantic reasoning for workflow deployment, resource allocation, workload adaptation, and migration across distributed computing environments. To demonstrate these capabilities, the ontology is populated with a hierarchical federated learning workload deployed over a representative smart grid infrastructure.

The AI pipeline consists of the following sequential stages: (1) preprocessing tasks that ingest, filter, and aggregate local raw grid measurements; (2) local training tasks that compute localized model updates and expose them via Eclipse Dataspace Connector to (3) a federated aggregation job; and (4) a global validation job evaluates the performance of the final global model; the final step is conditioned by the accuracy obtained, if the model satisfies the criteria the system moves to the (5.A) grid inference job; otherwise it goes into a (5.B) retraining loop. This workflow is shown in Figure 4.

*Corresponding author. E-mail address: tudor.cioara @cs.utcluj.ro, (T.Cioara).

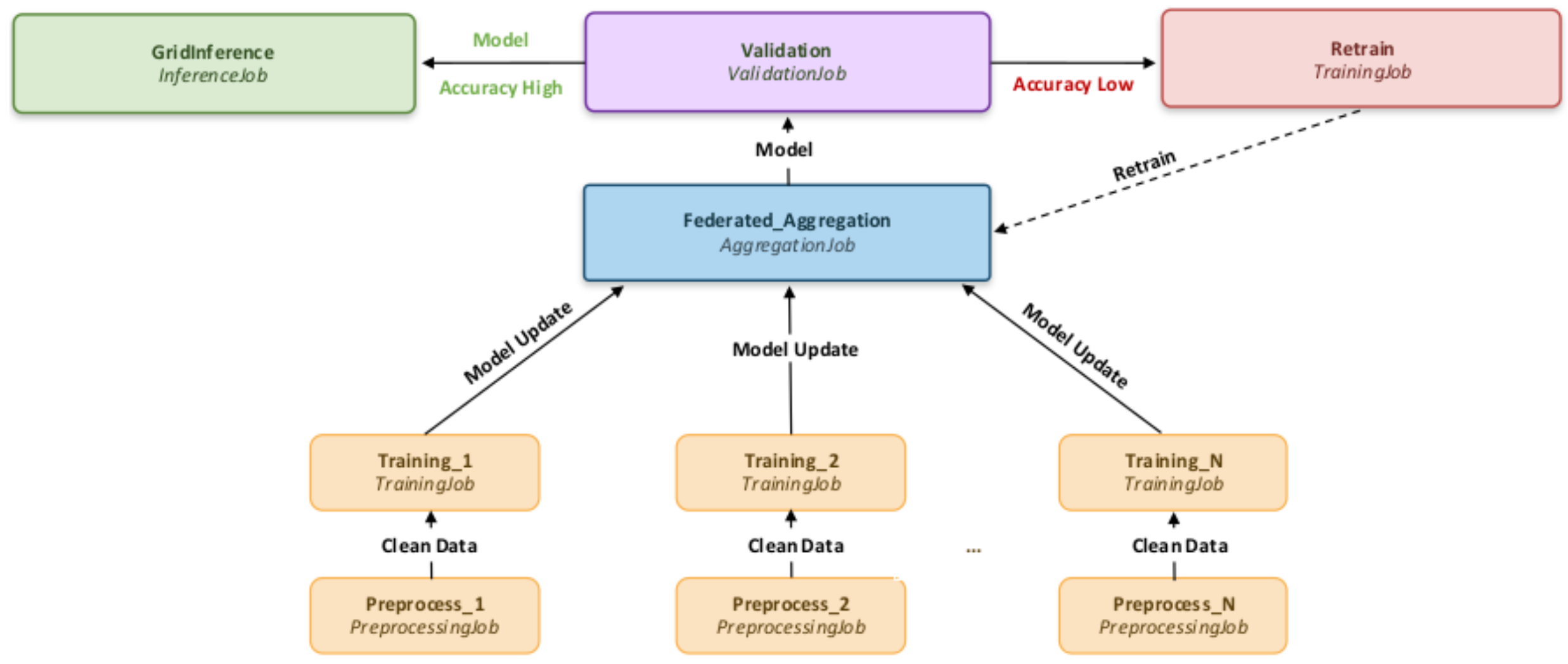

Figure 3. Federated AI pipeline

The AI Pipeline is generated and managed by the HEDGE-IoT orchestrator using the semantic information provided by DAI-ECC. The final workflow configuration depends on the raw measurement sources available in the target infrastructure. During workflow generation, the orchestrator discovers the available data sources through the ontology and instantiates the corresponding preprocessing and local training jobs for each source. The execution order is defined by the *hasNextStep* relationships, while data dependencies are represented through the *produces* and *consumes* properties. Therefore, the ontology describes the workflow as an abstract execution blueprint, whose concrete configuration is automatically derived considering the characteristics of the available computational infrastructure. Once the workflow has been instantiated, the orchestrator assigns each AI job to a computational node using a First Fit placement strategy based on the available infrastructure resources.

To showcase the adaptability of the proposed ontology, we generated over 100 random infrastructures of varying scale, each assigned with equal ~33% probability to one of three size classes: small (8-16 nodes), medium(17-30), large (31-40 nodes). After establishing the number of nodes within the infrastructure, each node is independently assigned a type (edge probability 60%, fog probability 30%, and cloud probability 10%) and a hardware profile where each dimension (number of cores, free memory, free storage, and GPU) is drawn random from a predefined array of candidates. Finally, each node independently has a probability of already running a pre-existing, migratable workload that consumes a part of the CPU and RAM resources. These workloads are represented in the ontology and may be migrated during deployment reasoning to enable the placement of new AI jobs (see Figure 4).

*Corresponding author. E-mail address: tudor.cioara @cs.utcluj.ro, (T.Cioara).

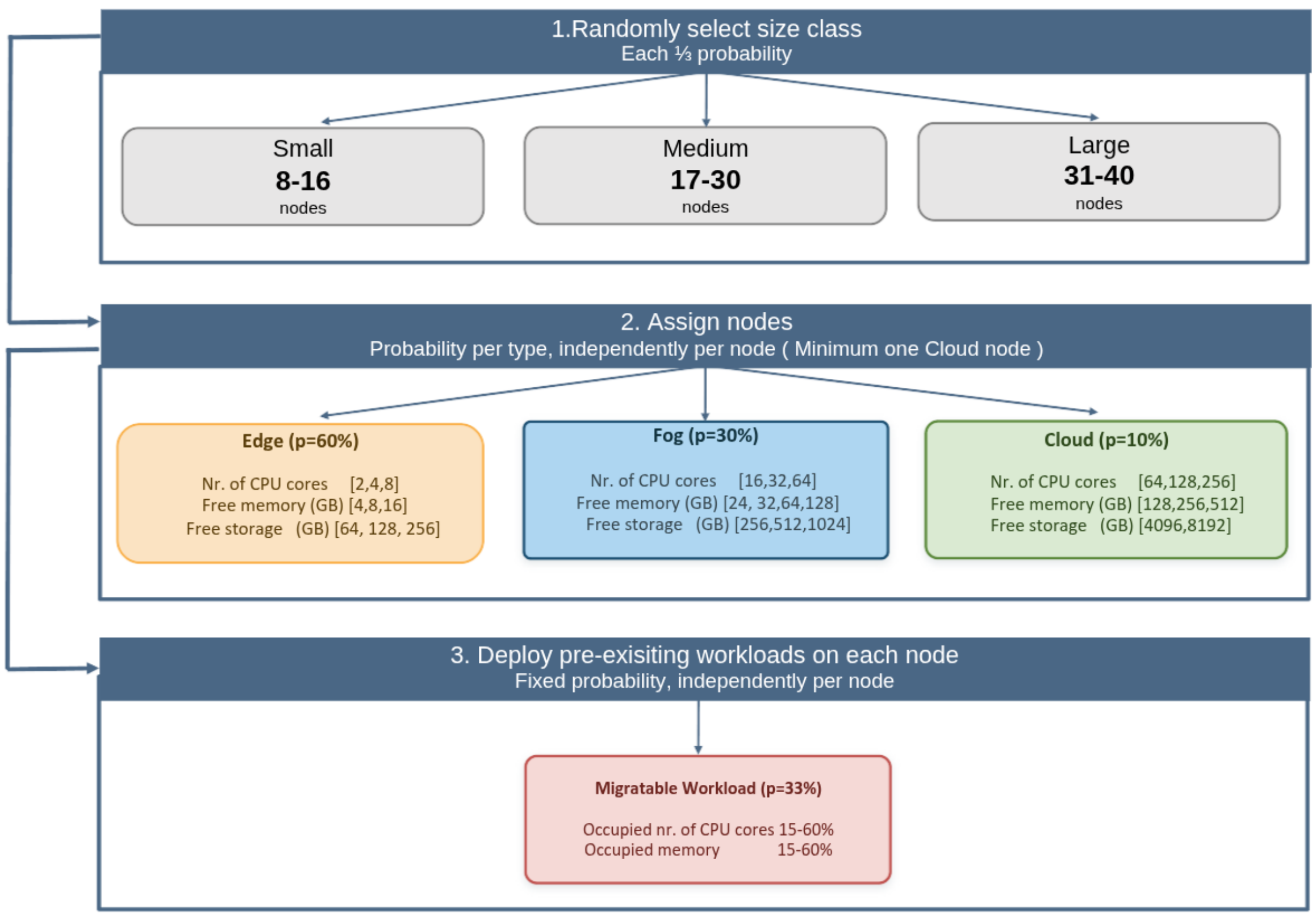

Figure 4. Randomly generated edge-fog-cloud infrastructure used to evaluate the DAI-ECC ontology.

For the infrastructures generated we instantiated the workflow with 5, 10, 15, 20, and 25 data sources and attempted deployment, where data sources count doesn't exceed the number of edge and fog nodes available to host them. The workflow stages are annotated with resource requirements such as CPU, memory, storage, and GPU, while nodes are described by their computational profiles. This unified semantic representation enabled the orchestrator to perform resource-aware placement by matching workflow requirements with node capabilities. AI jobs were processed considering the workflow sequence and assigned to the first computational node meeting their resource constraints. When no suitable node was available, the orchestrator attempted to migrate pre-existing workloads to alternative nodes, thereby releasing sufficient resources for the pending deployment.

The orchestrator interacts with the ontology through SPARQL queries to obtain the information required for deployment reasoning. The following SPARQL queries are used to retrieve the node state for placement:

- "What is the current available capacity of each computational node?"
- "Which AI jobs remain to be deployed, and what are their execution requirements?"
- "Which jobs are currently running on node X, and what resources do they consume?"

Each deployment attempt results in one of three possible outcomes: (i) successful deployment without workload migration, (ii) successful deployment requiring workload migration, or (iii) deployment failure due to insufficient available resources.

Figure 5 illustrates a successful deployment in which all AI jobs were placed on available computational nodes without requiring workload migration. The deployment was obtained by matching the resource requirements of each AI job with the available capabilities of the infrastructure nodes represented in DAI-ECC.

*Corresponding author. E-mail address: tudor.cioara @cs.utcluj.ro, (T.Cioara).

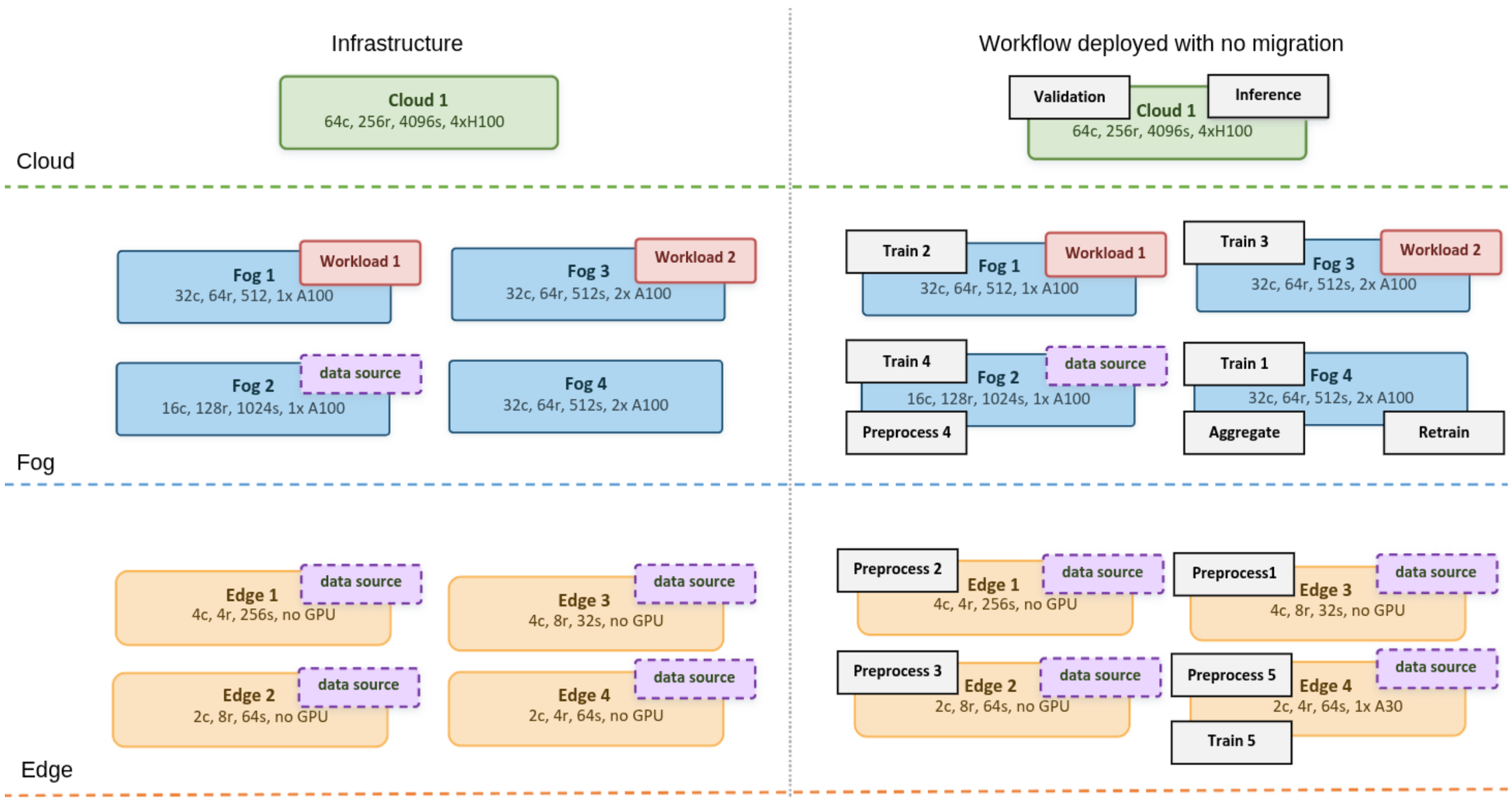

Figure 5. Small infrastructure, 5 data sources, success with no migration scenario

Figure 6 shows a deployment scenario in which the initial placement failed because no node satisfied the resource requirements of a pending AI job. By reasoning over the ontology, the orchestrator identified a migratable workload occupying resources on a suitable node, relocated it to an alternative host, and successfully completed the deployment.

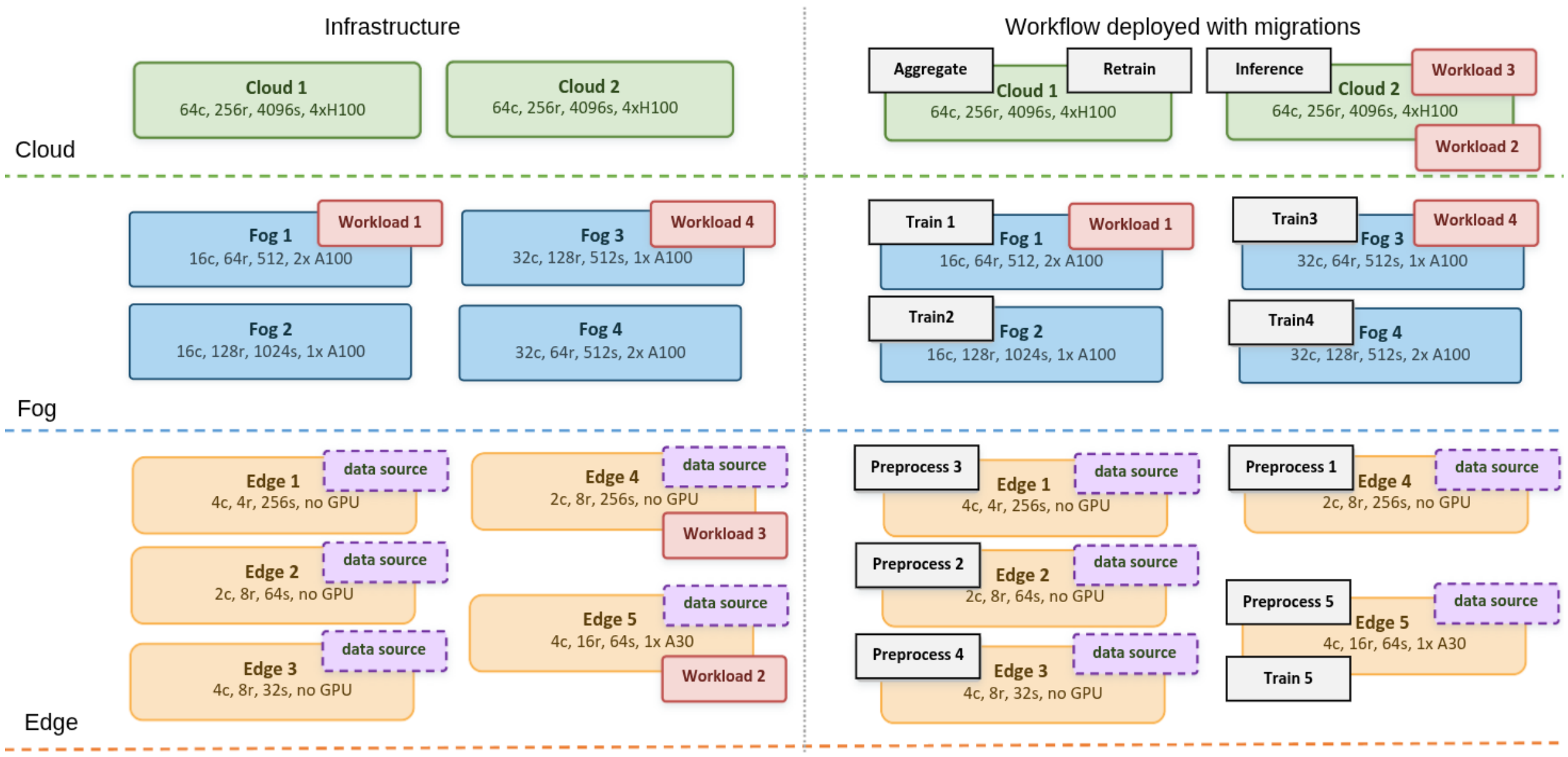

Figure 6. Small infrastructure, 5 data sources, success with migration scenario

Figure 7 illustrates a deployment failure. Although workload migration was considered, no feasible reallocation could free sufficient computational resources to satisfy the remaining deployment constraints. Consequently, the ontology-driven reasoning correctly concluded that the workflow could not be deployed on the available infrastructure.

*Corresponding author. E-mail address: tudor.cioara @cs.utcluj.ro, (T.Cioara).

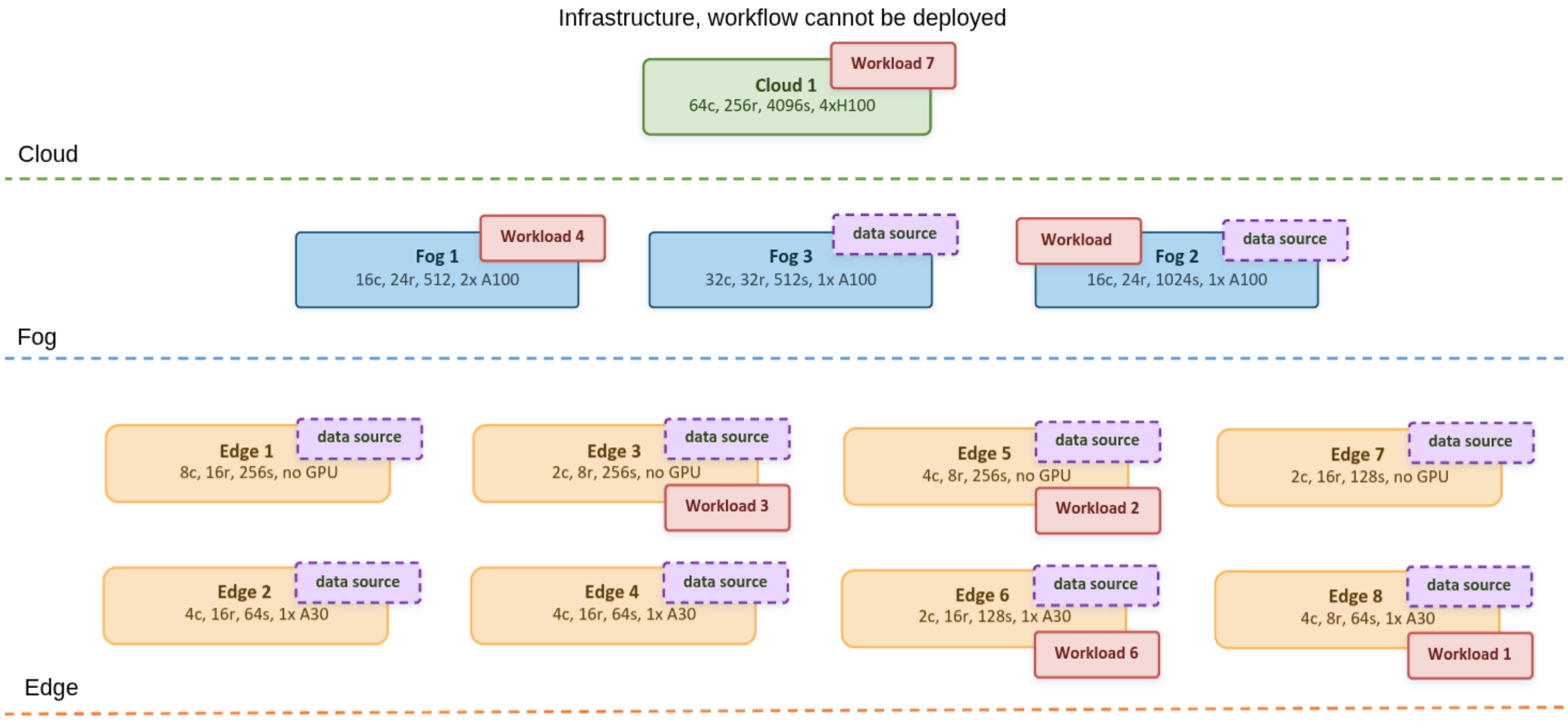

Figure 7. Small infrastructure, 10 data sources, fail scenario

Together, these scenarios demonstrate that DAI-ECC supports reasoning over heterogeneous computational infrastructures by combining descriptions of AI workflows, resource requirements, and infrastructure capabilities within a unified semantic model. This enables the orchestrator to distinguish between directly deployable workflows, deployments requiring workload migration, and infeasible deployments.

Table 9 summarizes the deployment outcomes and the average orchestration decision time for AI workflows of increasing size. Across all workflow configurations, the ontology enabled successful deployment in most cases, with most deployments requiring no workload migration. Only a small proportion of deployments required migration to satisfy resource constraints, while deployment failures occurred only when the available infrastructure could not accommodate the workflow, even after considering workload relocation. As expected, the average orchestration decision time increased gradually with the number of data sources, reflecting the larger search space associated with more complex workflows. Nevertheless, decision times remained below 80 ms for all evaluated configurations, demonstrating that ontology-driven reasoning introduces only a small computational overhead.

Table 9: Infrastructure configurations and results

| **Deployment outcome** | **# Data Sources** | | | | |
|---|---|---|---|---|---|
| | **5** | **10** | **15** | **20** | **25** |
| **Successful (no migrations)** | 86.7% | 100% | 93.3% | 86.7% | 84% |
| **Successful (with migrations)** | 6.7% | 0% | 6.7% | 10% | 0% |
| **Failed** | 6.7% | 0% | 0% | 3.3% | 16% |
| **Average decision time (ms)** | 57.5 | 62.4 | 63.5 | 69.4 | 77.3 |

The deployment experiments demonstrate the practical applicability of DAI-ECC in supporting AI workflows orchestration. Also, we have evaluated the ontology against the competency questions defined during the requirements specification phase to verify that it satisfies its intended functional requirements. The competency questions are divided into two categories according to the functionality required from the ontology: retrieval and reasoning. For retrieval the competency questions are translated to queries that navigate the knowledge to return facts that are reachable via property paths, while the second group is reasoning that classify instance by evaluating rules on constraints across jobs and infrastructure jointly. Table 10 shows three representative questions in detail.

*Corresponding author. E-mail address: tudor.cioara @cs.utcluj.ro, (T.Cioara).

Table 10: Relevant competency questions and associated implementation over DAI-ECC

<table>
<tr><th>Competency Question</th><th>Rules</th></tr>
<tr><td>

**CQ6. For the federated aggregation, which data are consumed, from which nodes do they originate, and what is the communication latency?**

**Purpose.** This competency question evaluates the ontology's ability to represent data dependencies and communication relationships in distributed AI workflows.

**Validation.** A SPARQL query traverses the semantic relationships linking aggregation jobs, exchanged model updates, producing training jobs, computational nodes, and network links to retrieve the origin of each consumed model update together with the corresponding communication latency.

**Result.** The query successfully returns every consumed model update, its originating computational node, and the associated communication latency, demonstrating that DAI-ECC integrates workflow semantics with infrastructure connectivity.

</td><td>

```
PREFIX s4mind: <https://saref.etsi.org/saref4mind/>
PREFIX rdfs:   <http://www.w3.org/2000/01/rdf-schema#>

SELECT ?aggJob ?dataObject ?producingJob ?originNode ?link ?latencyMs
WHERE {
  ?aggJob a s4mind:AggregationJob .

  OPTIONAL { ?aggJob s4mind:runOn ?aggNode }

  OPTIONAL { ?aggJob s4mind:usesDataConnectionPoint ?cp }

  OPTIONAL { ?cp s4mind:exposesDataObject ?dataObject }

  OPTIONAL { ?dataObject s4mind:isProducedBy ?producingJob }

  OPTIONAL { ?producingJob s4mind:runOn ?originNode }

  OPTIONAL {
    ?link s4mind:connectsNode ?originNode , ?aggNode ;
          s4mind:hasLatency ?latencyMs .
  }
}
ORDER BY ?aggJob ?originNode
```

</td></tr>
<tr><td>

**CQ8. Which computational nodes can host each AI job in the workflow?**

**Purpose.** This competency question evaluates the ontology's ability to support resource-aware deployment by determining whether each AI job can be assigned to at least one computational node satisfying its execution requirements.

**Validation.** Deployment feasibility is assessed by comparing the resource requirements associated with each AI job (e.g., CPU cores, memory, storage, accelerator type, accelerator count, and accelerator memory) with the corresponding hardware capabilities of candidate computational nodes. A single SPARQL query retrieves all feasible deployment nodes for each AI job, and flags the jobs that cannot be placed on the available infrastructure

**Result.** The query successfully identifies the deployable computational nodes for each AI job and detects jobs that cannot be deployed on the available infrastructure, demonstrating that DAI-ECC supports resource-aware deployment analysis across heterogeneous edge-fog-cloud environments.

</td><td>

```
PREFIX s4mind: <https://saref.etsi.org/saref4mind/>
PREFIX rdfs: <http://www.w3.org/2000/01/rdf-schema#>

SELECT ?pipeline ?job
       (COUNT(DISTINCT ?node) AS ?numFeasibleNodes)
       (GROUP_CONCAT(DISTINCT STRAFTER(STR(?node), "saref4mind/");
                     SEPARATOR=", ") AS ?feasibleNodes)
WHERE {

  ?pipeline a s4mind:AIPipeline ;
            s4mind:hasStep ?job .

  OPTIONAL {

    ?node a s4mind:ComputeNode .

    OPTIONAL { ?job s4mind:requiredCpuCores ?reqCpu }
    OPTIONAL { ?node s4mind:hasCpuCores ?nodeCpu }
    FILTER(!BOUND(?reqCpu) ||
           (BOUND(?nodeCpu) && ?nodeCpu >= ?reqCpu))

    OPTIONAL { ?job s4mind:requiredRamGB ?reqRam }
    OPTIONAL { ?node s4mind:hasMemory ?nodeRam }
    FILTER(!BOUND(?reqRam) ||
           (BOUND(?nodeRam) && ?nodeRam >= ?reqRam))

    OPTIONAL { ?job s4mind:requiredAcceleratorType ?reqAcc }
    OPTIONAL { ?node s4mind:hasAcceleratorType ?nodeAcc }
    FILTER(!BOUND(?reqAcc) ||
           ?reqAcc = ?nodeAcc)

    OPTIONAL { ?job s4mind:requiredAcceleratorCount ?reqAccCnt }
    OPTIONAL { ?node s4mind:hasAcceleratorCount ?nodeAccCnt }
    FILTER(!BOUND(?reqAccCnt) ||
           (BOUND(?nodeAccCnt) && ?nodeAccCnt >= ?reqAccCnt))

    OPTIONAL { ?job s4mind:requiredAcceleratorMemory ?reqAccMem }
    OPTIONAL { ?node s4mind:hasAcceleratorMemory ?nodeAccMem }
    FILTER(!BOUND(?reqAccMem) ||
           (BOUND(?nodeAccMem) && ?nodeAccMem >= ?reqAccMem))
  }
}
GROUP BY ?pipeline ?job
ORDER BY ?pipeline ?job
```

</td></tr>
</table>

*Corresponding author. E-mail address: tudor.cioara @cs.utcluj.ro, (T.Cioara).

| Competency Question | Rules |
| --- | --- |
| **CQ9. Which AI jobs require workload migration before deployment?**<br>**Purpose.** This competency question evaluates the ontology's ability to detect deployments that become invalid following changes in the available computational resources.<br>**Validation.** Deployment validity is assessed by re-evaluating the resource requirements of each deployed AI job against the current capabilities of its hosting computational node. A pipeline is flagged for reconfiguration if at least one of its AI jobs can no longer be supported by its assigned node.<br>**Result.** The query successfully identifies the AI pipelines and workflow stages requiring reconfiguration, demonstrating that DAI-ECC supports continuous deployment validation and provides the semantic basis for adaptive orchestration and workload migration. | |

```sparql
PREFIX s4mind: <https://saref.etsi.org/saref4mind/>
PREFIX rdfs:   <http://www.w3.org/2000/01/rdf-schema#>

SELECT DISTINCT ?pipeline ?job ?assignedNode
WHERE {
  ?pipeline a s4mind:AIPipeline ;
            s4mind:hasStep ?job .
  ?job s4mind:runOn ?assignedNode .

  OPTIONAL { ?job s4mind:requiredCpuCores ?reqCpu }
  OPTIONAL { ?assignedNode s4mind:hasCpuCores ?nodeCpu }

  OPTIONAL { ?job s4mind:requiredRamGB ?reqRam }
  OPTIONAL { ?assignedNode s4mind:hasMemory ?nodeRam }

  OPTIONAL { ?job s4mind:requiredAcceleratorCount ?reqAccCnt }
  OPTIONAL { ?assignedNode s4mind:hasAcceleratorCount ?nodeAccCnt }

  OPTIONAL { ?job s4mind:requiredAcceleratorMemory ?reqAccMem }
  OPTIONAL { ?assignedNode s4mind:hasAcceleratorMemory ?nodeAccMem }

  FILTER (
    (BOUND(?reqCpu) &&
      (!BOUND(?nodeCpu) || ?nodeCpu < ?reqCpu)) ||

    (BOUND(?reqRam) &&
      (!BOUND(?nodeRam) || ?nodeRam < ?reqRam)) ||

    (BOUND(?reqAccCnt) &&
      (!BOUND(?nodeAccCnt) || ?nodeAccCnt < ?reqAccCnt)) ||

    (BOUND(?reqAccMem) &&
      (!BOUND(?nodeAccMem) || ?nodeAccMem < ?reqAccMem))
  )
}
```

Table 11 summarizes competency questions used to validate DAI-ECC. For each competency question, the table identifies the ontology concepts and instances involved, the validation approach, and the obtained result. The competency questions are grouped according to the ontology capability they assess, namely information retrieval through SPARQL queries and semantic reasoning for deployment and orchestration decisions. All evaluated competency questions were successfully answered, demonstrating that the ontology supports both querying and reasoning over distributed AI workflows and computational infrastructures.

Table 11: Competency questions evaluation

| CQ | Ontology concepts | Ontology instances | Validation | Result |
| --- | --- | --- | --- | --- |
| **CQ1** | hasStep | AIPipeline, AIJob | SPARQL retrieval | ✓returns the jobs of the pipeline and their types |
| **CQ2** | Produces, hasDataObjectType | AIJob, DataObject | SPARQL retrieval | ✓returns each producing stage, its output and type |
| **CQ3** | hasStep. hasNextStep | AIPipeline, AIJob, | SPARQL retrieval | ✓returns the execution order and identifies the branch points |
| **CQ4** | exposesDataObject, isProducedBy, runOn, hasDataObjectType | EclipseDataSpaceConnectionPoint, DataObject, AIJob, ComputeNode | SPARQL retrieval | ✓returns the data crossing the boundary with producing job and origin node |
| **CQ5** | exposesDataObject, hasDataObjectType | EclipseDataSpaceConnectionPoint, DataObject | SPARQL retrieval | ✓returns no results, confirming no raw dataset is exposed across the boundary |

*Corresponding author. E-mail address: tudor.cioara @cs.utcluj.ro, (T.Cioara).

| CQ6 | usesDataConnectionPoint, exposesDataObject, isProducedBy, runOn, connectsNode, hasLatency | AggregationJob, EclipseDataSpaceConnectionPoint, DataObject, ComputeNode, NetworkLink | SPARQL retrieval | ✓returns each update, its origin node, and the link latency |
|---|---|---|---|---|
| **CQ7** | requiredCpuCores/RamGB/AcceleratorType/Count/Memory, hasCpuCores/Memory/AcceleratorType/Count/Memory maxAllowedLatency, connectsNode, hasLatency | AIJob, ComputeNode, NetworkLink | SPARQL reasoning | ✓returns each update, job and its feasible nodes meeting hardware and network constraints |
| **CQ8** | hasStep, requiredCpuCores/RamGB/AcceleratorType/Count/Memory, hasCpuCores/Memory/AcceleratorType/Count/Memory | AIPipeline, AIJob, ComputeNode | SPARQL reasoning | ✓returns each job, and deployable nodes for them |
| **CQ9** | hasStep, runOn, requiredAcceleratorCount, hasAcceleratorCount | AIPipeline, AIJob, ComputeNode; | SPARQL reasoning | ✓returns the pipeline and each job that requires migration |

# 6. Conclusions

In this paper we present DAI-ECC, a SAREF-compliant ontology for representing distributed AI workflows across the edge–fog–cloud continuum. We extended the ETSI SAREF4SYST ontology with concepts for AI pipelines, executable AI jobs, computational resources, deployment constraints, and communication relationships. In this way DAI-ECC provides a unified semantic model that bridges AI workflow semantics with heterogeneous computing infrastructures. The ontology enables semantic interoperability, automated reasoning, and resource-aware orchestration while remaining fully aligned with the ETSI SAREF ecosystem and its ontology engineering principles.

The ontology was validated by assessing its logical consistency and inferencing capabilities using standard reasoning tools, as well as through competency-question-based evaluation. All competency questions were successfully answered, demonstrating that the ontology supports both SPARQL querying and semantic reasoning over distributed AI workflows and computational infrastructures. In addition, the applicability of the proposed model was demonstrated through a real-world case study derived from the HEDGE-IoT project, focusing on the orchestration of federated learning energy services deployed across heterogeneous edge-fog-cloud infrastructures. Finally, it paves the way towards a semantic foundation for distributed AI orchestration that can be extended to a wide range of edge-fog-cloud applications. Its compatibility with the SAREF ecosystem positions it as a promising basis for future standardization efforts, in line with the European Commission's ICT Rolling Plan, which identifies the evolution of SAREF toward edge computing, federated machine learning, and generative AI as a strategic priority.

## Acknowledgments

This research received funding from the European Union's Horizon Europe research and innovation program under the Grant Agreements number 101136216 (Hedge-IoT). Views and opinions expressed are, however, those of the author(s) only and do not necessarily reflect those of the European Union or the European Climate, Infrastructure, and Environment Executive Agency. Neither the European Union nor the granting authority can be held responsible for them.

*Corresponding author. E-mail address: tudor.cioara @cs.utcluj.ro, (T.Cioara).

*Corresponding author. E-mail address: tudor.cioara @cs.utcluj.ro, (T.Cioara).

*Corresponding author. E-mail address: tudor.cioara @cs.utcluj.ro, (T.Cioara).

*Corresponding author. E-mail address: tudor.cioara @cs.utcluj.ro, (T.Cioara).